\documentclass[10pt,twocolumn,letterpaper]{article}

\usepackage{wacv}
\usepackage{times}
\usepackage{epsfig}
\usepackage{graphicx}
\usepackage{amsmath}
\usepackage{amssymb}
\usepackage{booktabs}
\usepackage{enumerate}
\usepackage{enumitem}
\usepackage{soul}
\usepackage{booktabs}
\usepackage{multirow}
\usepackage{algorithm}
\usepackage{algorithmic}
\usepackage{makecell}
\usepackage{subcaption}

\def\wacvPaperID{518} 

\wacvalgorithmstrack   

\wacvfinalcopy 

\ifwacvfinal
\usepackage[breaklinks=true,bookmarks=false]{hyperref}
\else
\usepackage[pagebackref=true,breaklinks=true,colorlinks,bookmarks=false]{hyperref}
\fi

\begin{document}

\title{Revisiting Training-free NAS Metrics: An Efficient Training-based Method}

\author{Taojiannan Yang$^{1}$\thanks{Work done during an internship at Bytedance Inc.}, Linjie Yang$^2$, Xiaojie Jin$^2$, Chen Chen$^1$ \\
$^1$Center for Research in Computer Vision, University of Central Florida \quad $^2$Bytedance Inc. \\
{\tt\small taoyang1122@knights.ucf.edu} \quad {\tt\small \{linjie.yang, jinxiaojie\}@bytedance.com} \\ {\tt\small chen.chen@crcv.ucf.edu}
}


\maketitle
\thispagestyle{empty}

\begin{abstract}
   Recent neural architecture search (NAS) works proposed training-free metrics to rank networks which largely reduced the search cost in NAS. In this paper, we revisit these training-free metrics and find that: (1) the number of parameters (\#Param), which is the most straightforward training-free metric, is overlooked in previous works but is surprisingly effective, (2) recent training-free metrics largely rely on the \#Param information to rank networks. Our experiments show that the performance of recent training-free metrics drops dramatically when the \#Param information is not available. Motivated by these observations, we argue that metrics less correlated with the \#Param are desired to provide additional information for NAS. We propose a light-weight training-based metric which has a weak correlation with the \#Param while achieving better performance than training-free metrics at a lower search cost. Specifically, on DARTS search space, our method completes searching directly on ImageNet in only 2.6 GPU hours and achieves a top-1/top-5 error rate of 24.1\%/7.1\%, which is competitive among state-of-the-art NAS methods. Codes are available at \url{https://github.com/taoyang1122/Revisit_TrainingFree_NAS}
\end{abstract}

\section{Introduction}
\label{sec:intro}

Neural Architecture Search (NAS) \cite{nasrl, regularizedevolution, progressivenas, enas, wu2019fbnet, liu2018darts, nasnet} is becoming an important technique in designing efficient and effective deep neural networks. Its effectiveness has been demonstrated in various computer vision tasks such as classification~\cite{enas,wu2019fbnet,nasnet}, object detection~\cite{chen2019detnas, tan2020efficientdet} and semantic segmentation~\cite{chen2019fasterseg, liu2019autodeeplab}. Early NAS methods \cite{nasrl, regularizedevolution, tan2019mnasnet} leverage reinforcement learning or evolutionary algorithm to search networks. But this process is extremely expensive because they need to train thousands of candidate networks. Following works \cite{liu2018darts, pdarts, pcdarts} alleviate this problem using differentiable search with candidate networks sampled from a supernet. During training, the network parameters and architecture parameters are optimized alternatively. However, training supernet can still be very slow and the accuracy of sub-networks in the supernet has a poor correlation with their ground truth accuracy \cite{yu2019evaluating}. To further reduce the search cost, training-free metrics \cite{naswot, tenas, zerocostproxy} are proposed to rank the candidate networks without any training process. These metrics are largely inspired by the pruning methods \cite{lee2018snip, grasp, synflow} and theoretical findings in deep neural networks \cite{ntk1, ntk2, linearregion, linearregion2}. They aim to rank the networks from different aspects of the networks' properties such as trainability and expressivity. These metrics achieve competitive results with previous NAS methods at a much smaller search cost.

However, these works overlooked a straightforward training-free metric, the \textit{number of parameters} (\#Param) in a network, which is even faster to compute than those training-free metrics. Our experiments show that \#Param is surprisingly good on NAS-Bench-101 \cite{nas101} and NAS-Bench-201 \cite{nas201}. We further discover that these training-free metrics have a very high correlation with \#Param (details in Sec. \ref{subsec:revisit}), which indicates that a large portion of their ranking ability may come from the correlation with \#Param. 
To validate our conjecture, we design systematic experiments to remove the impact of \#Param. The results show that without the \#Param information, recent training-free metrics \cite{tenas, naswot} do not achieve a good performance. 

Motivated by the above discovery, our objective is to develop a metric that has a weak correlation with \#Param while still being effective so that it can provide additional information on estimating the performance of a network.
Intuitively, a network's final performance is indicated by the structure (\eg, \#Param, \#Layers), weight initialization, and the dynamics during training (\eg, loss, gradients). We believe that metrics arise from the training dynamics should be weakly correlated with \#Param. Angle metric is a training dynamic which is first proposed in \cite{layerrotation} to indicate the network's generalization ability. It is defined as the angle between the vectorized network weights before and after training. We find that the angle metric at the final fully-connected (FC) layer has a high correlation with the accuracy but a low correlation with the number of parameters. This indicates that it can provide additional information other than \#Param on estimating the network's performance. To reduce the computation for model training, we propose an extremely light-weight training scheme with a small proxy dataset which is thousands times faster than traditional training. Our experiments show that such a short-training scheme already yields effective angle metrics. Besides the angle metric, we also leverage the training loss as a second metric, which achieves better performance without additional computation.
To summarize, we make the following contributions.
\begin{enumerate}[leftmargin=*]
    \item We revisit recent training-free metrics and reveal how they achieve good performance on the evaluated benchmarks. Although training-free metrics claim to rank networks by estimating the model's capacity and convergence speed, our experiments show that they achieve good performance mainly because they have high correlation with \#Param, and \#Param happens to be a good metric on the evaluated NAS benchmarks. \ul{Their functionality is in fact similar to \#Param while being unnecessarily complicated}.
    
    \item Motivated by our discovery, we propose a training-based metric which provides orthogonal information to \#Param on ranking networks. Our method achieves competitive performance with training-free methods on popular NAS benchmarks, and the performance will be significantly better when the \#Param information is not helpful. Our search cost is even smaller than training-free metrics.
    \item Our findings raise the necessity to design new search spaces where \#Param does not dominate the model performance to better evaluate the effectiveness of a NAS metric and understand how it works. Our results also inspire future works to design metrics that provide orthogonal information to \#Param because \#Param may not be a good metric in many cases (\eg, MLP vs. CNN).
\end{enumerate}

\section{Related work}
\label{sec:related_work}
\noindent\textbf{Neural architecture search (NAS).} NAS is proposed to search network structures automatically for a given task instead of time-consuming manual design. Early works \cite{nasrl, regularizedevolution, tan2019mnasnet, liu2018hierarchicalnas} leverage reinforcement learning or evolutionary algorithms to explore architectures. The controller will generate some networks and the network performance will be used as feedback information to update the controller. However, training a large amount of networks is very expensive, costing thousands of GPU days. Following works accelerate NAS algorithms by weight-sharing in a supernet. ENAS \cite{enas} proposes to share the weights among candidate networks so that they can be trained simultaneously. DARTS \cite{liu2018darts} concatenates all candidate operations into a supernet and each operation is assigned an architecture parameter denoting its importance. During training, the architecture parameters and weight parameters are optimized alternatively. Another kind of weight-sharing method is one-shot NAS \cite{one-shot1, brock2017smash, guo2020singlepath}, where a supernet is trained with sub-networks stochastically sampled in each iteration. However, recent  study~\cite{yu2019evaluating} shows that the network performance via weight-sharing has a poor correlation with its actual performance.

\noindent\textbf{Training-free NAS.} To further speedup the search process, recent works \cite{naswot, tenas, zerocostproxy} propose to predict network performance without training. \cite{zerocostproxy} evaluates the effectivenss of different pruning-at-initialization criteria \cite{grasp, synflow, lee2018snip} for NAS. NASWOT \cite{naswot} leverages the number of linear regions \cite{linearregion} to rank different networks. TE-NAS \cite{tenas} further combines linear regions with neural tangent kernel (NTK) \cite{ntk1} to rank a network by its expressivity and trainability. However, \cite{mok2022demystifying} shows that NTK-based metrics are unstable across different search spaces and initializations. In this work, we further reveal that the effectiveness of training-free metrics (Linear Region and NTK) mainly come from the high correlation with \#Param, and \#Param is a good metric on the evaluated benchmarks.

\section{Methodology}
\label{sec:method}
In Sec. \ref{subsec:revisit}, we first revisit several existing training-free metrics and \#Param. We demonstrate that \#Param is an effective search metric on NAS-Bench-101 and NAS-Bench-201, and that existing training-free metrics rely on \#Param to achieve high performance. Then we introduce our light-weight training-based metric and short-training strategy in Sec. \ref{subsec:angle_metric} and Sec. \ref{subsec:short_training}, respectively.

\subsection{Revisiting training-free metrics}
\label{subsec:revisit}



The number of \textbf{linear regions (LR)} is used in \cite{naswot, tenas} to rank networks at initialization. Linear region is a well-studied theoretical criteria \cite{linearregion, linearregion2} to indicate the learning capacity of a network. It is defined as how many regions a network could split the input space into. A larger number of linear regions indicates that the network has higher performance. The number of LR is estimated differently in TE-NAS \cite{tenas} and NASWOT \cite{naswot}. TE-NAS calculates LR by forwarding a batch of samples to the network and count how many samples have different activation patterns, while NASWOT feeds a batch of samples to the network and compute the Hamming distance between different activation patterns. The Hamming distance between these activation patterns is used to define a kernel matrix $\mathbf{K}$. The ranking metric is defined as the determinant of $\mathbf{K}$. To distinguish these two metrics, we denote the LR estimated by TE-NAS and NASWOT as LR1 and LR2, respectively.

TE-NAS further leverages the \textbf{neural tangent kernel (NTK)} to score networks. \cite{ntk1, ntk2} point out that the network's convergence speed is determined by the condition number of NTK. Intuitively, a faster convergence speed indicates that the network has a higher performance. So the condition number of NTK can be used to rank networks. Note that in \cite{tenas}, NTK is negtively correlated with the accuracy while in this paper we use negative NTK to make it positive.

These theoretical indicators describe a network's property from different perspectives. However, the most naive indicator to describe a network would be the \textbf{number of parameters (\#Param)}. Intuitively, a larger model tends to have better performance. This makes us wonder whether the number of parameters is a good training-free metric? The answer is yes. In Tab. \ref{tab:param_acc}, we show the comparison of \#Param and training-free metrics on NAS-Bench-101 \cite{nas101} and NAS-Bench-201 \cite{nas201}. We evaluate these metrics based on random search. Specifically, we randomly sample 100 networks from the search space and use the metrics to select the best one. We run each experiment 5 times and report mean accuracy and standard deviation. Surprisingly, the results show that \#Param achieves comparable performance with other training-free metrics on different datasets. 


\ul{The good performance of \#Param further motivates us to investigate whether these training-free metrics are correlated with \#Param.} We compute the Kendall rank correlation coefficient (Kendall's Tau) \cite{kendall1938new} between different training-free metrics and \#Param on NAS-Bench-101 (10000 networks) and NAS-Bench-201 (15625 networks) in Tab. \ref{tab:param_kt}. As a reference, the correlation between LR1 and LR2 is 0.56 on NAS-Bench-201. Note that they are the same metric just estimated differently, thus a correlation of 0.56 is high. The results show that all these training-free metrics have high correlations with \#Param, especially the two linear region metrics. This is intuitively plausible because the number of linear regions is upper bounded by $2^{\#activations}$, while the number of activation units is highly correlated with the number of parameters. These results imply that the ranking ability of these training-free metrics may mainly come from the high correlation with \#Param. In Sec.  \ref{sec:empirical_study}, we validate this conjecture by evaluating training-free metrics on networks of the same number of parameters. Their performance drops dramatically in this situation.

\textbf{What are the drawbacks of metrics having high correlation with \#Param?} Firstly, these training-free metrics claim to rank networks by estimating the model's capacity and convergence, but their functionality is in fact similar to \#Param while being unnecessarily complicated. Secondly, \#Param is not always a good metric. In the scenarios where the \#Param is not helpful (\eg, MLP vs. CNN, Residual vs. Plain structure, networks with similar \#Param as in Sec. \ref{sec:empirical_study}), the performance of such metrics will drop dramatically.

Motivated by these observations, we explore a new type of metric in this work, which is weakly correlated with the number of parameters while providing additional information on estimating the performance of the neural networks. Our proposed metric is introduced in the following sections.

\begin{table}[t]
  \centering
  \captionsetup{font=small}
  \caption{Comparison of \#Param and training-free metrics on NAS-Bench-101 and NAS-Bench-201. Each experiment is repeated 5 times and mean accuracy and standard deviation are reported.}
  \resizebox{\linewidth}{!}{
  \begin{tabular}{l|c|ccc}
    \toprule
    \multirow{2}{*}{Metrics} & NAS-Bench-101 & \multicolumn{3}{c}{NAS-Bench-201} \\
      & CIFAR-10 & CIFAR-10 & CIFAR-100 & ImageNet16-120 \\
    \hline
    \#Param & 92.6(1.3) & \textbf{93.2(0.5)} & \textbf{70.1(0.8)} & 41.6(4.1) \\
    LR1 & 91.6(0.9) & 92.3(1.1) & 66.2(5.0) & 43.1(2.5) \\
    NTK & 91.2(0.9) & 91.9(1.7) & 66.6(4.3) & 41.4(4.9) \\
    LR2 & \textbf{92.8(1.2)} & 92.6(0.9) & 69.3(1.4) & \textbf{43.3(2.9)} \\
    \bottomrule
  \end{tabular}
  }
  \label{tab:param_acc}
\end{table}

\begin{table}[t]
    \centering
    \captionsetup{font=small}
    \caption{Correlation (Kendall's Tau) of different training-free metrics with the number of parameters (\#Param).}
    \label{tab:param_kt}
    \small{
    \begin{tabular}{l|ccc}
    \toprule
         Correlation with \#Param & LR1 & NTK & LR2  \\
         \hline
         NAS-Bench-101 & 0.46 & 0.36 & 0.62 \\
         NAS-Bench-201 & 0.39 & 0.30 & 0.56 \\
         \bottomrule
    \end{tabular}
    }
\end{table}

\subsection{Angle metric}
\label{subsec:angle_metric}
Since existing training-free metrics all have a high correlation with the number of parameters based on the observations in Sec. \ref{subsec:revisit}, we shift our attention to the training dynamics. \textbf{Angle metric} is a training dynamic which is first proposed in~\cite{layerrotation} to indicate the generalization ability of a network and later used in~\cite{anglenas, randomlabelnas} as a metric to rank candidate networks in NAS. Considering all the weights of a network as a one-dimensional vector, angle metric is defined as the angle between the weight vectors before and after training. Specifically, let $\boldsymbol W_0$ denote the weights of a network $\boldsymbol N$ at initialization, and $\boldsymbol W_t$ denote the weights after training. Then the angle metric is defined as
\begin{equation}
   \theta(\boldsymbol N) = \arccos \left(\frac{\boldsymbol{W_0} \cdot \boldsymbol{W_t}}{\Vert \boldsymbol{W_0} \Vert_2  \Vert \boldsymbol{W_t} \Vert_2}\right),
\end{equation}
where $\boldsymbol{W_0} \cdot \boldsymbol{W_t}$ is the inner product of $\boldsymbol W_0$ and $\boldsymbol W_t$. \cite{randomlabelnas} shows that the angle metric is positively correlated with a network's final performance.

However, we find that the angle metric behaves differently at different network stages. Specifically, the angle metric computed with the weights from the feature extraction layers is positively correlated with the network's final accuracy, while the angle metric computed with the weights of the prediction layer (the final fully-connected layer) is negatively correlated with the performance. In most NAS search spaces~\cite{nas101, nas201, liu2018darts}, the feature extraction stage is mainly constructed by a stack of network modules. We denote the angle metric of the feature extraction stage $\theta_{feat}$ and the angle metric of the prediction layer $\theta_{pred}$ for brevity. 

In Tab .\ref{tab:angle_val1}, we demonstrate the impact of model parameters on above two variants of angle metrics through two kinds of network settings.
We randomly sample 50 networks with different sizes (setting 1) and the same size (setting 2) from NAS-Bench-201, and fully train them on CIFAR-10. Then we compute the Kendall's Tau of $\theta_{feat}$ and $\theta_{pred}$ for these two scenarios. In setting 1, it shows that $\theta_{feat}$ is positively correlated with the accuracy, which is consistent with \cite{anglenas, randomlabelnas}, but $\theta_{pred}$ is negatively correlated and has a higher correlation than $\theta_{feat}$.
However, in setting 2, the Kendall's Tau of $\theta_{feat}$ degrades dramatically to around 0, which means $\theta_{feat}$ fails to rank the networks without the \#Param information. But the Kendall's Tau of $\theta_{pred}$ degenerates less and is still able to rank the networks of the same number of parameters. Therefore, $\theta_{pred}$ is a metric with weak dependency on the number of parameters.


\begin{table}[t]
    \centering
    \captionsetup{font=small}
    \caption{Comparison of Kendall's Tau of $\theta_{feat}$ and $\theta_{pred}$ on 50 random networks with different sizes (different \#Param) or the same size (same \#Param), respectively.}
    \label{tab:angle_val1}
    \small
    \begin{tabular}{l|cc}
    \toprule
        Sampled Networks & $\theta_{feat}$ & $\theta_{pred}$ \\
        \hline
        diff. \#Param &  0.37 & -0.50 \\
        same \#Param &  -0.09 & -0.25 \\
    \bottomrule
    \end{tabular}
\end{table}

\subsection{Short-training scheme}
\label{subsec:short_training}
In Sec. \ref{subsec:angle_metric}, we show $\theta_{pred}$ is a good metric at ranking networks even without the \#Param information. However, fully training all candidate networks is too expensive in NAS. To alleviate this problem, we propose an extremely light-weight short-training scheme by using a small proxy dataset from the original target dataset. Specifically, we first randomly sample a sub-set of classes from the target dataset. Then for each sampled class, we randomly sample a small amount of images, generating a highly condensed proxy dataset. We train networks on the proxy dataset for a limited number of iterations. This training procedure is thousands times faster than fully training a network. 
We find our $\theta_{pred}$ metric is effective under such a compact setting in different search spaces and datasets.

Besides $\theta_{pred}$, we also use another training dynamic, the training loss, as an additional metric to evaluate networks. Note that training loss comes for free in our method. In Sec. \ref{sec:empirical_study}, we show that training loss also has weak correlation with the number of parameters. Combining training loss with $\theta_{pred}$ gives richer information on model performance without increasing the computational cost.

Since the scales of $\theta_{pred}$ and training loss are different, directly adding their values will cause one dominating the other. To avoid this problem, we first use these two metrics to rank networks respectively. Then we add their ranking index as the final ranking index of each network. Note that both $\theta_{pred}$ and training loss are negatively correlated with the model accuracy. For clarity, we take the negative value of the two metrics to make them positive in the following experiments. Since the proposed metric employs a short period of training, we name our NAS method combined with this metric as  Short-Training NAS (ST-NAS). A pipeline of ST-NAS based on random search is shown in Algorithm \ref{algo:pwnas}.

\begin{algorithm}[t]
\small
\captionsetup{font=small}
\caption{ST-NAS}
\begin{algorithmic}
\STATE \textbf{Input:} Number of candidate networks $N$. Search space $\mathcal{S}$. Target dataset $\mathcal{D}$. Training iterations $m$.
\STATE \textbf{Output:} Model with the highest rank.
\STATE $\triangleright$ Initialization
\STATE $\theta_{pred}$ = zeros($N$), $loss$ = zeros($N$)
\STATE sampler = RandomSampler()
\STATE Sample proxy dataset $\mathcal{\tilde{D}}$ from $\mathcal{D}$
\STATE $\triangleright$ Evaluate candidate networks
\FOR{i in $0,1,...,N-1$}
    \STATE network = sampler($\mathcal{S}$)
    \STATE $W_0$ = network.fc.weights
    \STATE Train the network for $m$ iterations with $\tilde{D}$.
    \STATE $loss$[i] = - compute\_loss(network, $\mathcal{\tilde{D}}$)
    \STATE $W_t$ = network.fc.weights
    \STATE $\theta_{pred}$[i] = - compute\_angle\_metric($W_0$, $W_t$)
\ENDFOR
\STATE $\triangleright$ Combine two metrics
\STATE $R_{\theta_{pred}}$ = get\_rankings($\theta_{pred}$)
\STATE $R_{loss}$ = get\_rankings($loss$)
\STATE $R$ = $R_{\theta_{pred}}$ + $R_{loss}$
\STATE max\_idx = model index with the highest rank in $R$
\STATE \textbf{return:} $\mathcal{S}$[max\_idx]
\end{algorithmic}
\label{algo:pwnas}
\end{algorithm}

\begin{figure*}[t]
  \centering
  \captionsetup{font=small}
  \includegraphics[width=0.9\linewidth]{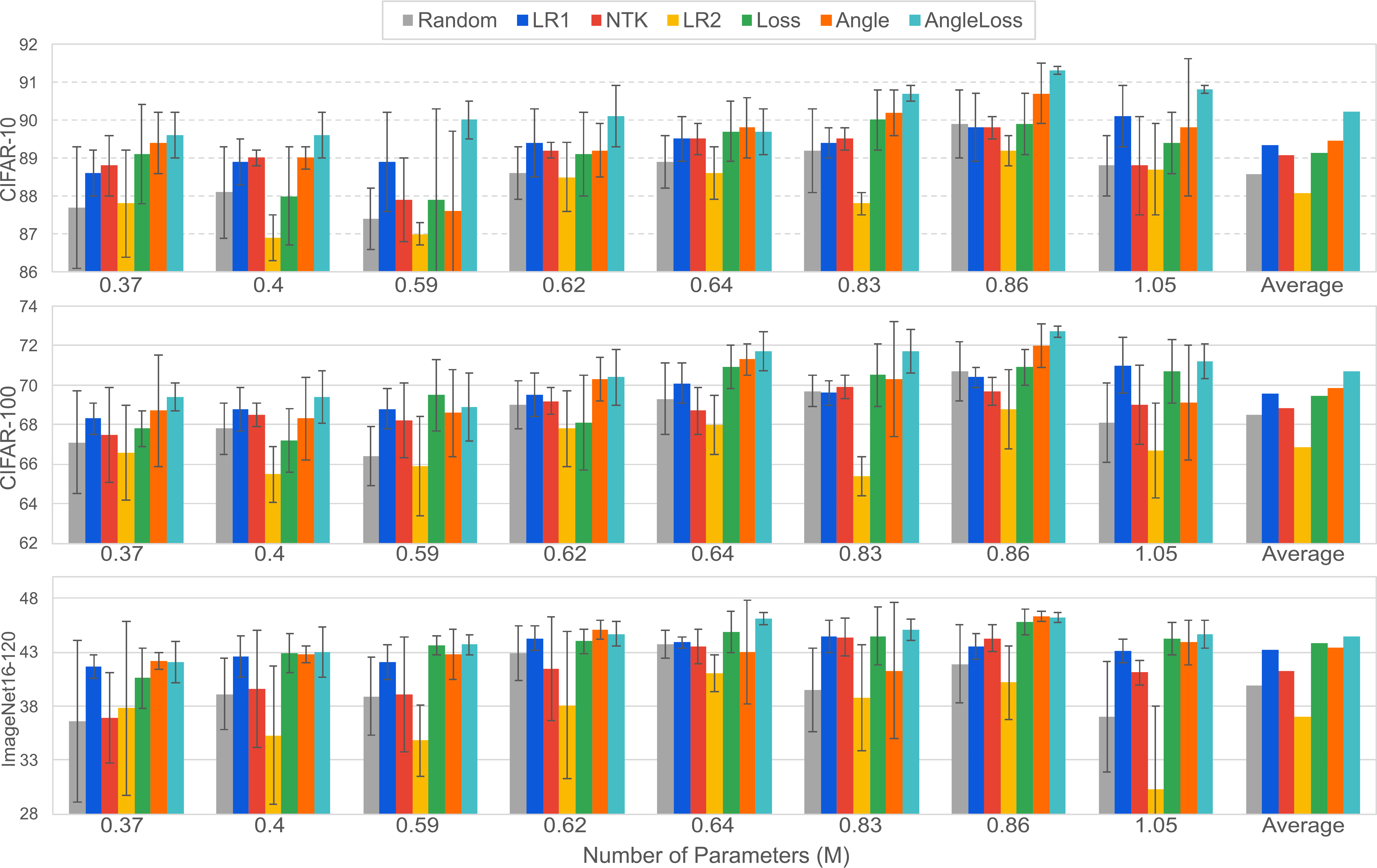}
   \caption{Test accuracy (\%) of different metrics when evaluated on networks of the same number of parameters. X-axis is the number of parameters (M) in each network group. Each experiment is repeated 5 times and the mean accuracy and standard deviation are reported.}
   \label{fig:paramgroup}
\end{figure*}

\section{Empirical study}
\label{sec:empirical_study}
As discussed in Sec. \ref{sec:method}, recent training-free metrics are highly correlated with the number of parameters, which implies their effectiveness comes from the high correlation with number of parameters. To further validate our claim, we thoroughly evaluate different training-free metrics and our metric on curated search spaces with the same number of parameters. This prevents metrics from leveraging the parameter information to evaluate networks. In the following sections, $Angle$ denotes searching with $\theta_{pred}$, $Loss$ denotes searching with training loss and $AngleLoss$ denotes searching with the combination of the two metrics.

We craft several search spaces based on NAS-Bench-201 \cite{nas201}. NAS-Bench-201 defines a cell-based search space. Each cell is represented as a densely-connected directed acyclic graph (DAG). Each cell has 4 nodes and 6 edges, where each edge represents an operation. There are 5 candidate operations, including \textit{zeroize, skip-connect, $1\times1$ conv, $3\times3$ conv, and $3\times3$ avg pooling}. Different models may have the same number of parameters but with different structures and performances. We choose 8 groups of models, and models in the same group has the same number of parameters, i.e. \{0.37, 0.40, 0.59, 0.62, 0.64, 0.83, 0.86, 1.05\} M, respectively. The number of networks in each group is \{1602, 540, 1602, 810, 180, 540,
180, 135\}, respectively. We evaluate the effectiveness of different metrics on each of these network groups. We compute the training-free metrics using the settings in the original papers \cite{tenas, naswot}. Our training scheme is detailed in Sec. \ref{subsec:short_training}. We randomly sample 10 classes and 10 images from each class. The network is trained for 50 iterations with a fixed learning rate of 0.2. Other settings follow those in NAS-Bench-201 \cite{nas201}. Note this is the default setting throughout our experiments if not specified.

\begin{table}[t]
    \centering
    \vspace{-0.5cm}
    \captionsetup{font=small}
    \caption{Kendall's Tau between our metrics and \#Param.}
    \small
    \begin{tabular}{l|ccc}
    \toprule
       Metrics  & Angle & Loss & AngleLoss \\
       \hline
       Correlation & 0.20 & -0.11 & 0.07 \\
       \bottomrule
    \end{tabular}
    \label{tab:angle_param_kt}
    \vspace{-0.2cm}
\end{table}

We compare the performance of previous training-free metrics and our metrics using random search. We randomly sample 100 networks from each network group and select the best-performing network per the metric. We also add a baseline which randomly selects a network from candidate networks. Each experiment is repeated 5 times and the mean accuracy and standard deviation are reported. As shown in Fig. \ref{fig:paramgroup}, LR2, which has the highest correlation with \#Param in Tab. \ref{tab:param_kt} and the best performance in Tab. \ref{tab:param_acc}, performs the worst in this scenario. It is even worse than the random baseline. Our $AngleLoss$ metric consistently outperforms training-free metrics on all the network groups on three datasets. In most cases, $AngleLoss$ is higher than training-free metrics by more than 1\%. 
We also show our metrics' Kendall's Tau with \#Param in Tab. \ref{tab:angle_param_kt}. As can be seen, the correlations are much lower than that of the training-free metrics in Tab. \ref{tab:param_kt}. Above experiment 
evidences that training-free metrics largely rely on the parameter information to rank networks, and that our metric is advantageous by having weak correlation with the number of parameters, providing additional useful information to estimate a network's performance.


\begin{table*}[t]
    \centering
    \captionsetup{font=small}
    \caption{Comparison of the test accuracy of different metrics on NAS-Bench-101 and NAS-Bench-201 based on random search ($N=100$). Each experiment is repeated 5 times to compute its mean and standard deviation.}
    \small
    \begin{tabular}{l|c|c|ccc}
    \toprule
     \multirow{2}{*}{Metrics} & \multirow{2}{*}{Search Cost (s)} & NAS-Bench-101  & \multicolumn{3}{c}{NAS-Bench-201} \\
          & & CIFAR-10 & CIFAR-10 & CIFAR-100 & ImageNet16-120 \\
        \hline
        \#Param & 3 & 92.58(1.26) & 93.21(0.49) & 70.15(0.83) & 41.58(4.07)  \\
        \hline
        LR1 & 60 & 91.98(1.31) & 92.30(1.07) & 66.23(4.96) & 43.12(2.52)  \\
        LR1+\#Param & 60 & 92.52(1.37) & 92.96(0.55) & 69.83(0.43) & 43.71(2.20)  \\
        \hline
        NTK & 181 & 91.23(1.11) & 91.94(1.70) & 66.63(4.29) & 41.38(4.88)  \\
        NTK+\#Param & 181 & 91.48(1.52) & 93.12(0.48) & 69.82(0.73) & 42.39(1.61)  \\
        \hline
        LR2 & 48 & 91.95(1.16) & 92.65(0.93) & 69.28(1.40) & 43.33(2.91) \\
        LR2+\#Param & 48 & 92.58(1.39) & 93.33(0.13) & 70.10(1.22) & 42.83(1.49) \\
        \hline
        AngleLoss & 437 & 92.86(0.77) & 84.65(5.88) & 58.06(0.40) & 28.08(0.31) \\
        AngleLoss+\#Param & 437 & \textbf{93.60(0.46)} & \textbf{93.46(0.59)} & \textbf{70.58(0.82)} & \textbf{43.74(1.48)} \\
        AngleLoss+LR2 & 462 & 93.47(0.47) & 93.08(0.66) & 69.62(0.59) & 43.43(1.62) \\
        \bottomrule
    \end{tabular}
    \label{tab:random_nas201}
\end{table*}

\section{Experiments}
\label{sec:experiments}
In Sec. \ref{subsec:results_nasbench201}, we first show the comparisons of training-free metrics and our metric on NAS-Bench-101 and NAS-Bench-201. We apply metrics to both random search method and pruning-based search method. Then we compare our metric with other methods on DARTS search space in Sec. \ref{subsec:results_darts}. Finally, we conduct ablation studies to show the impact of short-training hyperparameters.

\subsection{Results on NAS-Bench-101/201}
\label{subsec:results_nasbench201}

\textbf{Random search.}
We first evaluate different metrics based on random search. We randomly sample 100 networks from the search space and use different metrics to select the best one. We follow the default settings in \cite{naswot, tenas} to compute training-free metrics LR1, LR2, and NTK. Our training settings are the same as in Sec. \ref{sec:empirical_study}. We run each experiment 5 times and report the mean accuracy and standard deviation. The search cost is measured on a single GTX-1080Ti GPU.

The results are shown in Tab. \ref{tab:random_nas201}. We add \#Param as a baseline metric in Tab. \ref{tab:random_nas201}. It is shown that \#Param performs well on both NAS-Bench-101 and NAS-Bench-201. It is even slightly better than training-free metrics on CIFAR-10 and CIFAR-100. Note that \#Param is very easy to compute, with a search cost of only 3 seconds on 100 networks. The linear region based metrics (LR1 and LR2) are better and more stable than NTK. The performance of NTK is low and has a very large variance. Although both LR1 and LR2 are based on linear regions, LR2 is slightly better and more stable. Note the effectiveness of training-free metrics could be attributed to their high correlation with \#Param. 

Surprisingly, our metric $AngleLoss$ does not perform well on the overall search space of NAS-Bench-201, although we have demonstrated in Sec. \ref{sec:empirical_study} that it is significantly better than other training-free metrics in different network groups. By visualizing the searched network structures, we find that our $Angle$ metric could collapse to some trivial structures, where most of the connections are \textit{zeroize, skip-connect} or \textit{avg\_pooling}. Our conjecture is that in these trivial structures, the feature extraction layers are not learning anything meaningful, and the prediction layer is optimized towards random directions in each training iteration. So the weight vector of the prediction layer almost does not change after training, which means $Angle$ metric will give a high score to these structures. However, this problem could be easily resolved if we combine our metric with \#Param to avoid the structures with a small number of parameters. It can also be avoided when we use a pruning-based search method. In Tab. \ref{tab:random_nas201}, we see that our metric is significantly boosted by around 10\% when combined with \#Param, and it achieves higher performance than other training-free metrics. On NAS-Bench-101, we don't have the collapse problem because there are fewer trivial structures. We achieve significantly better performance than training-free metrics.

We also combine training-free metrics with \#Param. It shows that these training-free metrics can also slightly benefit from \#Param, but the improvement is marginal. \ul{Taking \#Param as the baseline, combined with training-free metrics will even degrade its performance on NAS-Bench-201 CIFAR-10 and CIFAR-100. However, our metric achieves consistent improvements upon \#Param on three datasets.} We also show that when combined with LR2, AngleLoss+LR2 improves upon LR2 on all datasets. These experiments demonstrate that our metric provides orthogonal information to \#Param and training-free metrics. They can be combined together to achieve better performance.

\begin{table*}[t]
    \centering
    \captionsetup{font=small}
    \caption{Comparison of the test accuracy on NAS-Bench-201 based on pruning-based search in \cite{tenas}. $^\dagger$ indicates the results are reproduced by us using the official released codes \cite{tenas_codes}. The search cost of our method and TE-NAS is measured on 1080Ti GPU while LGA is measured on Tesla A40 GPU. The \textbf{best} and \ul{second best} results are bold and underlined, respectively.}
    \small
    \begin{tabular}{l|c|ccc}
    \toprule
        Method & Search Cost (s) & CIFAR-10 & CIFAR-100 & ImageNet16-120 \\
        \hline
        RSPS \cite{rsps} & 8007 & 87.66(1.69) & 58.33(4.34) & 31.14(3.88) \\
        DARTS (1st) \cite{liu2018darts} & 10889 & 54.30(0.00) & 15.61(0.00) & 16.32(0.00) \\
        GDAS \cite{GDAS} & 28925 & 93.61(0.09) & 70.70(0.30) & 41.84(0.90) \\
        LGA \cite{mok2022demystifying} & 5400 & \textbf{93.94(N/A)} & \textbf{72.42(N/A)} & \textbf{45.17(N/A)} \\
        \hline
        TE-NAS \cite{tenas} & 1558 & \ul{93.90(0.47)} & \ul{71.24(0.56)} & 42.38(0.46) \\
        TE-NAS$^\dagger$ \cite{tenas} & \ul{682} & 93.20(0.29) & 70.44(1.34) & 42.34(0.63) \\
        \hline
        AngleLoss & \textbf{508} & 93.16(0.37) & 70.48(1.04) & 43.04(1.82) \\
        AngleLoss+\#Param & \textbf{508} & 93.36(0.26) & 70.87(0.41) & \ul{43.77(1.33)} \\
        \bottomrule
    \end{tabular}
    
    \label{tab:prune_nas201}
\end{table*}

\begin{table*}[t]
    \centering
    \captionsetup{font=small}
    \caption{Comparison with state-of-the-art on DARTS CIFAR-10. The \textbf{best} and \underline{second best} results are bold and underlined, respectively.}
    \small
    \begin{tabular}{l|cccc}
    \toprule
         Method & \makecell{Search Cost \\ (GPU days)} & Params (M) & Top-1 Acc (\%) & Search Method \\
        \hline
        NASNet-A \cite{nasnet} & 2000 & 3.3 & 97.35 & RL \\
        ENAS \cite{enas} & 0.5 & 4.6 & 97.11 & RL \\
        AmoebaNet-A \cite{regularizedevolution} & 3150 & 3.2 & 96.66 & evolution \\
        \hline
        Random baseline \cite{liu2018darts} & 4 & 3.2 & 96.71 & random \\
       DARTS (1st) \cite{liu2018darts} & 0.4 & 3.3 & 97.00 & gradient \\
        DARTS (2nd) \cite{liu2018darts} & 1.0 & 3.3 & 97.24 & gradient \\
        GDAS \cite{GDAS} & 0.17 & 2.5 & 97.18 & gradient \\
        P-DARTS \cite{pdarts} & 0.3 & 3.4 & \textbf{97.50} & gradient \\
        PC-DARTS \cite{pcdarts} & 0.1 & 3.6 & 97.43 & gradient \\
        SDARTS-ADV \cite{sdarts} & 1.3 & 3.3 & 97.39 & gradient \\
        TE-NAS \cite{tenas} & \textbf{0.05} & 3.8 & 97.37 & training-free \\
        \hline
        AngleLoss & \ul{0.09} & 3.2 & 97.37 & short-training \\
        AngleLoss+\#Param & \ul{0.09} & 3.2 & \ul{97.44} & short-training \\
        \bottomrule
    \end{tabular}
    
    \label{tab:darts_CIFAR}
\end{table*}

\textbf{Pruning-based search.}
We also apply our metric to pruning-based search used in TE-NAS \cite{tenas}. All the settings are the same as in Sec. \ref{sec:empirical_study}, except that we train the supernet for 100 iterations because it takes longer for the supernet to converge. Each experiment is repeated 5 times and the mean and standard deviation are reported.

We compare our method with TE-NAS in Tab. \ref{tab:prune_nas201}. The performances of some other NAS methods are cited from \cite{nas201} for reference. We report two results for TE-NAS, one is reported in the original paper \cite{tenas} and the other is reproduced by us using the official codes \cite{tenas_codes} since we cannot reproduce the results in the original paper using the default setting. The reproduced performance is lower while the search cost is also cheaper (we evaluate it on a 1080Ti GPU, which is the same as in TE-NAS). In Tab. \ref{tab:prune_nas201}, we can see that our short-training method is even faster than TE-NAS. This is because TE-NAS needs to compute two metrics (LR1 and NTK), and for each metric it repeats 3 times and takes the average value to have a better and stable performance. However, we only compute our metric once with an extremely short training scheme. 

Under the pruning-based search, our metric does not show the collapse problem as in random search. This is because pruning-based method starts from a supernet, which is definitely non-trivial. With a limited number of pruning steps, the network almost never reach a trivial structure with large numbers of empty operations. As shown in Tab. \ref{tab:prune_nas201}, the original results of TE-NAS are better than ours on CIFAR-10 and CIFAR-100, but the search cost is $3\times$ of ours. Our performance is comparable with the reproduced results of TE-NAS at a lower search cost. On ImageNet16-120, our metric is better than TE-NAS in both cases. We also combine our metric with \#Param with negligible additional search cost. It further improves our performance on all three datasets by $0.2\% - 0.7\%$.

\subsection{Results on DARTS search space}
\label{subsec:results_darts}
We apply our metric to the pruning-based search method used in TE-NAS \cite{tenas} for the following experiments. 

\textbf{Results on CIFAR-10.}
We first compare our metric with other methods on CIFAR-10 dataset. 
As shown in Tab. \ref{tab:darts_CIFAR}, our metric completes the search process in 0.09 days (\ie, 2.16 hours) on a single 1080Ti GPU. Different from the results on NAS-Bench-201, our search cost is higher than TE-NAS in this case. This is because TE-NAS uses a smaller batch-size to compute NTK on DARTS CIFAR-10, resulting in less computation. Nevertheless, our search cost is still much lower than other NAS methods. Our metric also achieves comparable performance with TE-NAS, but the searched network size is much smaller. When combined with \#Param, our metric again achieves a lower test error of 2.56\%, which is competitive with state-of-the-art methods.

\begin{table*}[t]
    \centering
    \captionsetup{font=small}
    \caption{Comparison with state-of-the-art NAS methods on DARTS search space ImageNet-1K dataset.}
    \large
    \resizebox{0.8\linewidth}{!}{
    \begin{tabular}{l|cccc|cc}
    \toprule
        Method & \makecell{Search Cost \\ (GPU days)} & Params (M) & Top-1 (\%) & Top-5 (\%) & Search Method & Search Dataset \\
        \hline
        NASNet-A \cite{nasnet} & 2000 & 5.3 & 74.0 & 91.6 & RL & \multirow{9}{*}{CIFAR-10} \\
        AmoebaNet-C \cite{regularizedevolution} & 3150 & 6.4 & 75.7 & 92.4 & evolution \\
        DARTS (2nd) \cite{liu2018darts} & 4.0 & 4.7 & 73.3 & 91.3 & gradient \\
        GDAS \cite{GDAS} & 0.21 & 5.3 & 74.0 & 91.5 & gradient \\
        P-DARTS \cite{pdarts} & 0.3 & 4.9 & 75.6 & 92.6 & gradient \\
        PC-DARTS \cite{pcdarts} & 0.1 & 5.3 & 74.9 & 92.2 & gradient \\
        TE-NAS \cite{tenas} & 0.05 & 6.3 & 73.8 & 91.7 & training-free \\
        AngleLoss & 0.09 & 4.7 & 75.3 & 92.5 & short-training \\
        AngleLoss+\#Param & 0.09 & 4.7 & 74.8 & 92.3 & short-training \\
        \hline
        ProxylessNAS \cite{progressivenas} & 8.3 & 7.1 & 75.1 & 92.5 & gradient & \multirow{5}{*}{ImageNet-1K} \\
        PC-DARTS \cite{pcdarts} & 3.8 & 5.3 & 74.8 & 92.7 & gradient \\
        TE-NAS \cite{tenas} & 0.17 & 5.4 & 75.5 & 92.5 & training-free \\
        AngleLoss & \textbf{0.11} & 4.8 & 74.5 & 91.9 & short-training \\
        AngleLoss+\#Param & \textbf{0.11} & 5.9 & \textbf{75.9} & \textbf{92.9} & short-training \\
        \bottomrule
    \end{tabular}
    }
    \label{tab:darts_imagenet}
\end{table*}

  

    

\begin{table*}[!t]
    \captionsetup{font=small}
    \caption{Ablation study of different training hyper-parameters on NAS-Bench-201 CIFAR-100.}
    \begin{subtable}{.5\linewidth}
      \centering
      \captionsetup{font=footnotesize}
        \caption{Number of training iterations.}
        \resizebox{\linewidth}{!}{
        \tiny
        \begin{tabular}{l|cccc}
            \#Iters & 10 & 25 & 50 & 75 \\
            \hline
            Cost (s) & 99 & 230 & 437 & 673 \\
            Acc (\%) & 70.22(1.08) & 70.33(0.91) & 70.58(0.82) & 70.37(0.57) \\
        \end{tabular}
        }
    \end{subtable}%
    \begin{subtable}{.5\linewidth}
      \centering
      \captionsetup{font=footnotesize}
        \caption{Number of sampled classes.}
        \resizebox{0.8\linewidth}{!}{
        \tiny
        \begin{tabular}{l|ccc}
            \#Classes & 5 & 10 & 20  \\
            \hline
            Cost (s) & 332 & 437 & 641 \\
            Acc (\%) & 70.02(0.74) & 70.58(0.82) & 70.30(0.74) \\
        \end{tabular}
        }
    \end{subtable}
    \begin{subtable}{.5\linewidth}
      \centering
      \captionsetup{font=footnotesize}
        \caption{Network initialization.}
        \resizebox{\linewidth}{!}{
        \tiny
        \begin{tabular}{l|ccc}
            Init. & Kaiming\_uniform & Kaiming\_normal & Xavier\_uniform  \\
            \hline
            Cost (s) & 437 & 437 & 437 \\
            Acc (\%) & 70.58(0.82) & 70.40(0.70) & 70.25(1.00) \\
        \end{tabular}
        }
    \end{subtable}
    \begin{subtable}{.5\linewidth}
      \centering
      \captionsetup{font=footnotesize}
        \caption{Number of sampled images.}
        \resizebox{0.8\linewidth}{!}{
        \tiny
        \begin{tabular}{l|ccc}
            \#Images & 5 & 10 & 20  \\
            \hline
            Cost (s) & 347 & 437 & 627 \\
            Acc (\%) & 70.26(1.08) & 70.58(0.82) & 70.28(0.97) \\
        \end{tabular}
        }
    \end{subtable}
    \label{tab:ablation}
    \vspace{-0.2cm}
\end{table*}

\textbf{Results on ImageNet-1K.}
We compare our metric with state-of-the-art NAS methods on ImageNet-1K \cite{imagenet} in Tab. \ref{tab:darts_imagenet}. Our short-training setting is the same as in CIFAR-10. For evaluation, we follow \cite{tenas} to stack the network with 14 cells and the initial number of channel is 48. In the top half of Tab. \ref{tab:darts_imagenet}, the networks are searched on CIFAR-10 and then evaluated on ImageNet-1K. We can see that our metric is competitive with state-of-the-art NAS methods with a much lower search cost. Compared to TE-NAS, our performance is significantly better and the network size is much smaller. The bottom half of Tab. \ref{tab:darts_imagenet} shows the results with different methods searched directly on ImageNet-1K. Pruning-based search with our metric completes in only 0.11 GPU days (\ie, 2.64 GPU hours), which is even faster than TE-NAS. Our metric is more than $30\times$ faster than the other NAS methods. The performance of our metric alone is slightly lower than other methods but with a smaller model size. When combined with \#Param, the performance of our metric is largely improved and reaches a competitive top-1/top-5 error rate of 24.1\%/7.1\%, outperforming listed differentiable and training-free methods. Note that our search cost is also significantly lower than other methods.

\subsection{Ablation Study}
Here we study the impact of different hyper-parameters in our short-training scheme,  including number of training iterations, sampled classes, images per class and weight initialization methods. We conduct experiments on CIFAR-100. The results of different settings are shown in Tab. \ref{tab:ablation}. We use the random search method in Tab. \ref{subsec:results_nasbench201} as the baseline. We can see that longer training iterations tend to achieve better performance. This is because longer training iterations allow the network to converge better, which yields more informative angle metric and training loss. But even only 10 training iterations can achieve a decent performance. Increasing the number of classes does not always improve the performance. We speculate that although more classes could provide more information about the target dataset, it also makes the proxy dataset harder, which makes the network harder to converge in the limited iterations and yields less informative angle metric and training loss. Similarly, increasing the number of images does not guarantee better performance either. To achieve the optimal accuracy-efficiency trade-off, one may need to tune the training hyper-parameters. But the performance is not very sensitive to the hyper-parameters and it is feasible to tune hyper-parameters because our method is highly efficient.

\section{Conclusion}
\label{sec:conclusion}
We conduct a systematic study to explore the relationship between recent training-free metrics and \#Param. 
Our empirical study shows that recent training-free metrics works similarly to \#Param while being unnecessarily complicated. 
Motivated by this discovery, we propose a light-weight training-based metric which provides orthogonal information than \#Param on estimating model performance. Our method achieves competitive performance with state-of-the-art NAS methods, while being even faster than training-free metrics. On the search spaces where the \#Param information is not useful, the performance of training-free metrics drops dramatically while our method significantly outperforms them on different datasets. We hope our work could inspire future works to design new metrics which provide more parameter-independent information on estimating the network's performance. 

{\small
\bibliographystyle{ieee_fullname}
\bibliography{egbib}
}

\end{document}